# 8-Valent Fuzzy Logic for Iris Recognition and Biometry

N. Popescu-Bodorin[*], *Member*, V.E. Balas[**], *Senior Member*, and I.M. Motoc[*], *Student Member*, IEEE

[*]Artificial Intelligence & Computational Logic Lab., Math. & Comp. Sci. Dept., 'Spiru Haret' University, Bucharest, România
[**]Faculty of Engineering, 'Aurel Vlaicu' University, Arad, România
bodorin@ieee.org, balas@drbalas.ro, motoc@irisbiometrics.org

*Abstract*- **This paper shows that maintaining logical consistency of an iris recognition system is a matter of finding a suitable partitioning of the input space in enrollable and unenrollable pairs by negotiating the user comfort and the safety of the biometric system. In other words, consistent enrollment is mandatory in order to preserve system consistency. A fuzzy 3-valent disambiguated model of iris recognition is proposed and analyzed in terms of completeness, consistency, user comfort and biometric safety. It is also shown here that the fuzzy 3-valent model of iris recognition is hosted by an 8-valent Boolean algebra of modulo 8 integers that represents the computational formalization in which a biometric system (a software agent) can achieve the artificial understanding of iris recognition in a logically consistent manner.**

I. INTRODUCTION

Because the visual acuity of the human agent is doubled by its intelligence – both of them together ensuring an excellent quality in indentifying the (dis)similarity of iris images, the geometry that illustrates the binary decisions given by the human agent during a Turing test [11] of iris recognition is very simple (Fig. 1.a): it consists of one collection of crisp points (0 and 1) and one histogram that counts how many times a decision of unitary score (1 - for the case of similar irides) or a null decision (0 - for the pairs of non-similar irides) was given by the human agent. Still, the geometry that illustrates the fuzzy binary decisions given by a software agent ([6]-[8]) during a Turing test of iris recognition is not that simple: in this case, the fuzzy biometric decisions given by the software agent define (draw) a f-geometry [13] in which the intra- and inter-class score distributions could be a little bit confused (Fig. 1.c, Fig. 2.a, Fig. 2.b), or confused much stronger (Fig. 1.b, Fig. 1.c in [6], Fig. 10 in [4]), or not confused at all. (Fig. 1.b from here, and Fig. 4.a, Fig. 4.b, Fig. 4.c in [6]).

*A. Crisp / Fuzzy Iris Recognition*

In fact, Fig. 1.a illustrates that iris recognition is crisp for a human agent, and consequently, the recognition function R (as it is perceived by the human agent) is a crisp indicator of the imposter (0) and genuine (1) classes of iris pairs (P):

$$R(\cdot,\cdot): P \rightarrow \{0,1\},$$

In concordance with the terminology introduced in [13], the function R (Fig. 1.a) will be referred to as *the prototype recognition function* and it is a crisp concept. The goal of designing automated iris recognition systems is to find fuzzy approximations f-R for the prototype recognition function R, as close as possible to R. Such an approximation f-R will be further referred to as a *fuzzy recognition function*. The fuzzy approximations f-R obtained by applying automated iris recognition methods are of the same types as those presented in Fig. 1.b (an excellent approximation, [7]), Fig. 1.c, Fig. 2.a, Fig. 2.b (very good approximations, [8]), Fig. 10 in [4] (good approximation), Fig. 1.b - Fig. 1.c and Fig. 4.a - Fig. 4.c in [6] (good approximations), where the marks (good, very good, excellent) were given using as a reference the result obtained in an approach considered nowadays as being the "state of the art" in iris recognition (and marked here as "good approximation" [4]).

*B. Why Crisp, Why Fuzzy?*

In the case in which the recognition is made using artificial agents and good quality eye images, the fact that the approximations f-R depart from the prototype R (situation illustrated in Fig. 1.b - Fig. 1.b.c and Fig. 4.a - Fig. 4.c from [6] and in Fig. 10 from [4]) can not be caused by the lack of visual acuity of the system, but only by the less intelligent manner in which the system decides (understands) iris similarity or dissimilarity. Practically, the artificial agent fuzzifies the prototype R and the separation between genuine and imposter score distributions. More inadequate and unintelligent the image processing is, much confusion it introduces in the biometric decision model. There are two significant differences between the ways in which the human agent and software agent decide the similarity or dissimilarity of two iris images:
- Ordinary people are not aware of the numerical reality of an image but only of certain meanings 'decoded' accordingly to their experience from the chromatic variation captured in the image. For the human agent the iris image is not a numerical data but a set of complex knowledge about the iris texture and the image quality (given by the technical acquisition conditions and the posture in which the eye is captured). The similarity/dissimilarity decision given by the human agent for a pair of iris images is based on ad-hoc techniques of comparing two such sets of knowledge, techniques which are adaptive in relation with the pair of images analyzed.
- An artificial agent makes the biometric decision using only numerical support. From its point of view, the iris image is numerical data in the first place. Depending on the intelligence with which it is endowed, the artificial agent can extract (artificial) knowledge about the numerical data, which is usually referred to as 'features', and further encoded in a numeric format. For example, the binary iris code ([1] - [4], [6]) is a binary encoding of the features extracted from a uint8 (8-bit unsigned integer) iris image. The artificial agent performs the comparison of two iris images indirectly, by comparing encoded features of the two iris images.

In short, the human agent operates in a rich knowledge space, whereas an artificial agent usually encodes the actual knowledge space in a relatively poor, partial and often imprecise numeric data, in a manner very similar to lossy compression. This is why the fuzzification is almost inherent in the ordinary practice of automated iris recognition.

*C. The problem*

As it was described above, any simple Turing test of iris recognition undertaken by using good quality images [9] confirms that different or identical iris images are easily and correctly recognized by human agents and fuzzy recognized by software agents. The cause of this happening is that the same problem is represented (projected) in different spaces of knowledge, or in other words, as it is intuitively illustrated in Fig. 2, humans and software agents see the iris recognition from different perspectives. For the human agent 'genuine' and 'imposter' are crisp and disjoint concepts, whereas for the artificial agent they are fuzzy concepts which sometimes share a confusion zone. The problem is how to reconcile these two different views that humans and artificial agents have on iris recognition. The solution is to find a suitable defuzzification of the imposter and genuine score distributions which guarantees that the fuzzy (and consequently the crisp) concepts 'genuine' and 'imposter' are disjoint, the appartenence of a pair of irides to these fuzzy or crisp sets being, in this case, mutually exclusive events.

*D. Related Works*

The papers investigating logical aspects of iris recognition or logical aspects of biometry in general are indeed very few. The situation when a pair of irides ambiguously belongs to both imposter and genuine fuzzy sets is investigated in [8]. It is shown there that in such case, artificial understanding of iris recognition experimental data is logically inconsistent. It is the case of wolf-lamb pair discussed in [12].

It is not the first time when we say it, *what iris recognition really is* and *how different providers of biometric solutions compete to each other* are two very different things. Still, due to this competition a lot of commented experimental data was published ([5], for example), all of them together involuntarily proving that Equal Error Rate (EER) is a crisp concept only in theory. The negative aspect of this competition is the fact that a lot of resources were invested to minimize EER value without a preliminary proper investigation of the suitable means of doing that, but as illustrated in Fig. 1.a and Fig. 1.d, the real improvement of iris recognition technology depends on finding and accepting a major change of perspective which implicitly leads to EER minimization. The first steps in this direction were undertaken with very good results in [8] and [7] (see Fig. 2.a, Fig. 2.b) by defining and simulating Intelligent Iris Verifier (IIV) and IIV Distributed System (IIVDS). However, the artificial (automated) understanding of the experimental data obtained in these simulations proved to be quite a difficult but rewarding task. The last section of this paper shows that clarifying the logical model of iris recognition allowed us to define iris recognition theory and the problem of designing improved iris recognition systems as classical problems of system identification [10]. In this perspective, designing iris recognition systems means identifying possible variables ([10], type I.a structure identification problem), relevant variables ([10], type I.b structure identification problem), input-output relation as a collection of fuzzy if-then Sugeno rules ([10], type II.a structure identification problem), the partitioning of the premise space ([10], type II.b structure identification problem), and doing all of these accordingly to the results of a Turing test (Fig. 1.a) and in a logically consistent manner.

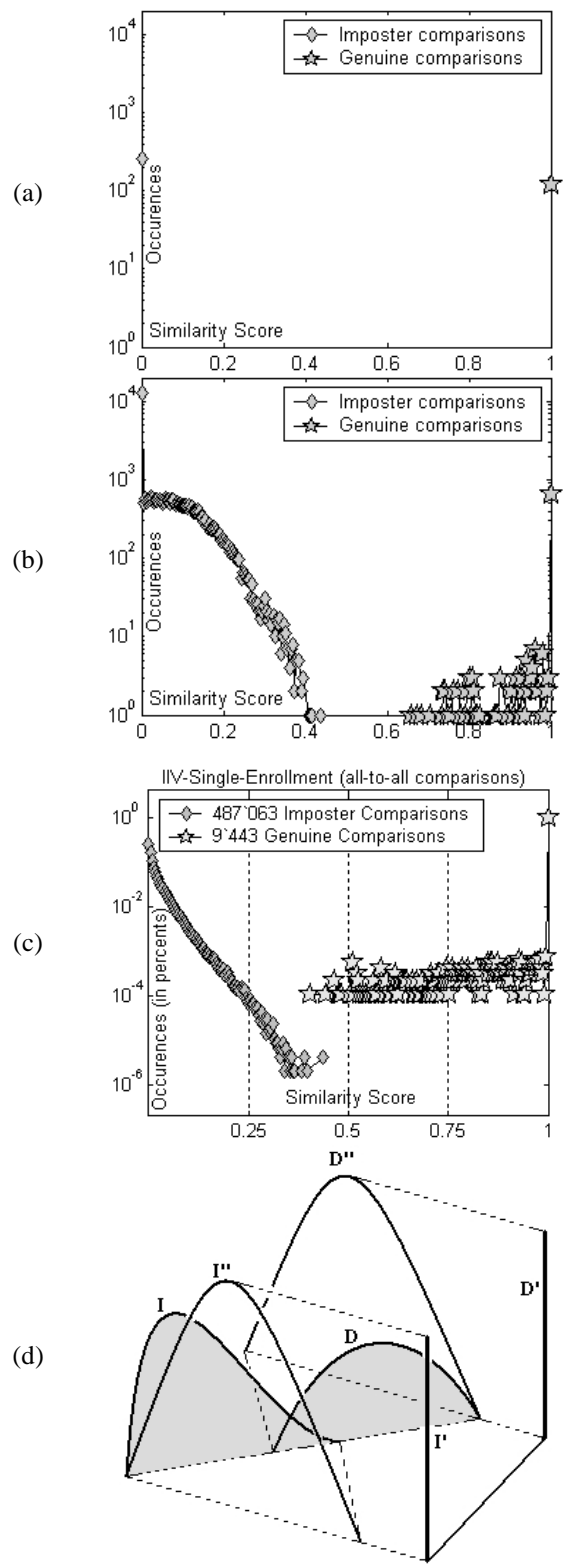

Fig. 1. (a) - The crisp geometry (0-1) of a biometric problem which is decidable in a consistent binary logic. (b) - The f-geometry of a biometric problem which is decidable in a fuzzy but still consistent binary logic. (c) The f-geometry of a biometric problem decidable in a fuzzy binary logic with very weak-confused fuzzy logical values. (d) - Viewing biometric decisions in iris recognition from different perspective: hypothetical genuine (I") and imposter (D") fuzzy clusters in real world as perceived by the human agent (I', D') and by the software agent (I, D).

The experimental data used in this paper is obtained in [8] and illustrated in Fig. 2.a and Fig. 2.b. The reason for using these data is that the imposter and genuine score distributions obtained in this case are much closer to the original recognition prototype function R (Fig. 1.a) than those obtained in other approaches.

II. A FUZZY 3-VALENT DISAMBIGUATED MODEL OF IRIS RECOGNITION

When 'genuine' and 'imposter' (pairs / comparisons) are fuzzy concepts / sets that share a (narrower or a wider) confusion zone, there are elements of vocabulary (irides / iris codes / digital identities) which ambiguously belong in both of them. Such a 2-valent fuzzy model of iris recognition is ambiguous and logically inconsistent [8]. Disambiguation is achieved by introducing a third fuzzy set, namely the fuzzy EER interval (f-EER) as a separator between the 'genuine' and 'imposter' fuzzy sets which in this way become disjoint, the appartenence of a recognition score to them being, in this case, mutually exclusive events.

*A. A Practical Example*

Disambiguation is a matter of system calibration and design (a type II.b structure identification problem, [10]) which must be carried out with respect to the desired FAR / FRR (False Accept / Reject Rate) specification. For instance, let us consider the requirement that a system must have very low FAR and FRR rates, 1E-10 to be more precise. Let us consider that we know a way to pessimistically estimate the FAR and FRR for scores where experimental data is not dense enough, or is completely missing, as POFA and POFR (Pessimistic Odds of False Accept / Reject). Hence, the recognition of identical / different irides will take place for all similarity scores t for which POFA(t)<1E-10, respectively for which POFR(t)<1E-10. Hence, the fuzzy EER interval (f-EER) is determined as (n,p), n=POFA$^{-1}$(1E-10), p=POFR$^{-1}$(1E-10)) which for the case considered is well approximated by the interval (0.3725, 0.55). Because FAR and POFA are both decreasing functions, and because FRR and POFR are both increasing functions (with respect to the similarity score), the comparisons scored in [0, 0.3725] and those scored in [0.55, 1] will be recognized by the system as being imposter / genuine comparisons, respectively. Hence, fuzzy decisions of the system can be encoded in three fuzzy values, I (Identical), D (Different) and O (Otherwise), corresponding to the partitioning ([10], type II.b structure identification problem) of the premise space (irides / iris codes / digital identities) as preimages of three intervals through a fuzzy recognition function f-R (Table 1, Table 2). Hence in a fuzzy 3-valent disambiguated model of iris recognition there are three kinds of iris code pairs: genuine, imposter and *undecidable*. The iris codes of an undecidable pair are, in fact, *unenrollable*, or else, the restriction of the system to the set of these pairs is logically inconsistent. This shows that in order to preserve logical consistency, each time when an iris code attempt to enroll in the system, one-to-all (one candidate iris code to all previously enrolled iris codes) comparisons are mandatory.

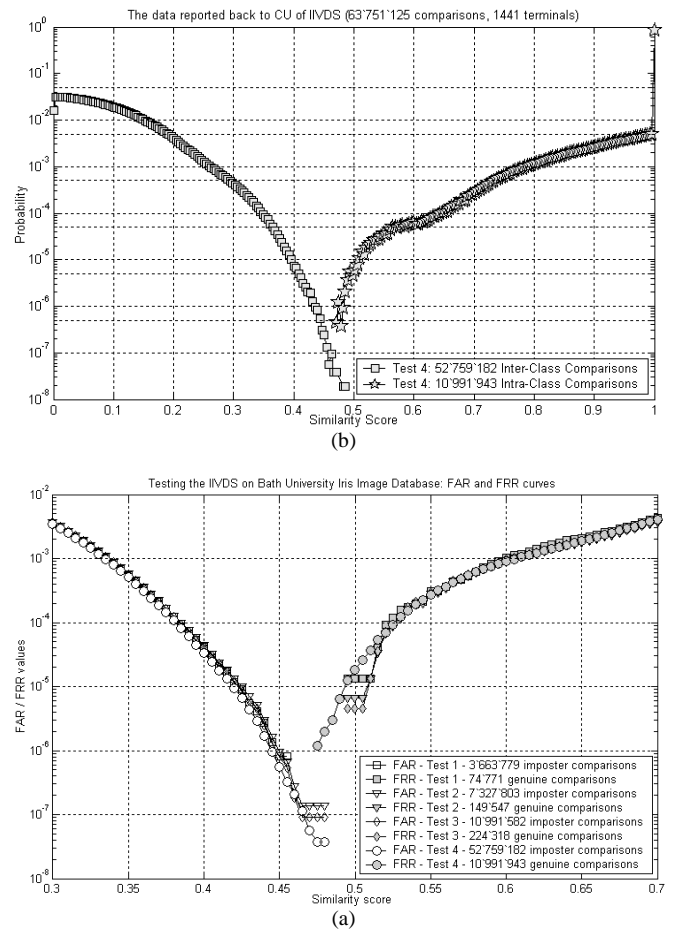

Fig. 3. (a) – The f-geometry of a biometric problem decidable in fuzzy binary logic with very weak confused logical values. (b) – Zoom within FAR-FRR curves corresponding to the iris recognition tests undertaken in [8]

TABLE 1: THE FUZZY 3-VALENT DISAMBIGUATED MODEL OF IRIS RECOGNITION APPLIED ON A PARTICULAR CASE [8]

| User attempt : Negative (1) or Positive (2) Claim | | System response quantified as recognition score: [0.55,1] – positive claim accepted, [0,0.3725] - positive claim rejected, [0.3725, 0.55] – uncertainty interval (f-EER). | | |
|---|---|---|---|---|
| | | Score | Decision / Interval | Decision / Meaning |
| 1 | I'm not X. Decline an enrolled digital identity (negative claim). | t, recognition / similarity score | I ≡ [0.55, 1] | False / Claim Rejected |
| | | | O ≡ (0.3725, 0.55) | Uncertain / Claim Repeat |
| | | | D ≡ [0, 0.3725] | True / Claim Accepted |
| 2 | I am X. Claim an enrolled digital identity (positive claim). | | I ≡ [0.55, 1] | True / Claim Accepted |
| | | | O ≡ (0.3725, 0.55) | Uncertain / Claim Repeat |
| | | | D ≡ [0, 0.3725] | False / Claim Rejected |

III. 8-VALENT ISOMORPHIC BOOLEAN ALGEBRAS WHICH CAN HOST A FUZZY 3-VALENT DISAMBIGUATED MODEL OF IRIS RECOGNITION

The Boolean algebra $I^3 = ((\emptyset, [0,n], (n,p), [p,1]), \cup, \cap, C)$ generated by the empty set, imposter, genuine and incertitude intervals with the reunion, intersection and complement, induces a formal logic that hosts fuzzy values I, O, and D. Still, this characterization is rather symbolic than computational. From a set of Boolean algebras isomorphic to $I^3$ we will choose one that can be expressed computationally (arithmetically). The candidates to choose from are illustrated in Table 5, where the empty fuzzy value (E) encodes (the empty set as the set of) the impossible states/decisions of a logically consistent biometric system: the states/decisions in which the system accepts both the positive and the negative claim regarding an iris code candidate and an identity is not observable in a logically consistent biometric system.

An algebra isomorphic to $I^3$ is $S^3 = ((E, I, O, D), \cup, \cap, C)$ - the Boolean algebra of strings (unsorted and with no repetition) generated through concatenation, intersection and complementary by the empty string E='' and the distinct characters 'I', 'O', 'D', corresponding to the modal values I, O and D.

The following function:
$$\forall s \in S^3: \text{str2bin}(s) =$$
$$= \text{va}('I' \in s) \cdot \vec{e_1} + \text{va}('O' \in s) \cdot \vec{e_2} + \text{va}('D' \in s) \cdot \vec{e_3}$$
is the isomorphism between the algebra of binary codes $B^3 = (\{0,1\}^3, \text{And}, \text{Or}, \text{Neg})$ and $S^3 = ((E, I, O, D), \cup, \cap, C)$, where:
- $\{\vec{e_1}, \vec{e_2}, \vec{e_3}\}$ is the canonical basis of $R^3$,
- function 'va' returns in binary digit the truth values of its argument,
- And, Or and Neg are the bit-wise logical operators.

In its turn, the Boolean algebra $B^3 = (\{0,1\}^3, \text{And}, \text{Or}, \text{Neg})$ is isomorphic to the Boolean algebra $V^3 = (\{0,1\}^3, \oplus, *, !)$ of the vectors on the unit cube in $R^3$, (Fig. 4.c), generated by the vectors of the canonical basis of $R^3$, where:
- '+', '-' are the sum and the difference of two vectors,
- '*' extracts the common (dependent) part of two vectors with respect to the canonical basis,
- $\oplus$ is defined by the relation $a \oplus b = a + b - a*b$,
- '!a' is the difference from vector 'a' to the main diagonal of the unit cube.

The following function:
$$\forall s \in B^3: \text{bin2oct}(s) = \sum_{k=0}^{2} s(k+1) \cdot 2^k$$
transforms the algebra $B^3 = (\{0,1\}^3, \text{And}, \text{Or}, \text{Neg})$ in the algebra of modulo 8 integers denoted $(Z_8, P, S, N)$ where:
- $N(a)$ is the complement of 'a' relative to 7,

TABLE 2: BIOMETRIC DECISION IN A FUZZY 3-VALENT DISAMBIGUATED MODEL OF IRIS RECOGNITION

| | | | |
|---|---|---|---|
| I | [0.55, 1] | Genuine pairs | **False Accept Rate:**<br>FAR(0.55) ≈ POFA(0.55) = 1E-10.<br>**True Accept Safety:**<br>1−FAR(0.55) ≈ 1−POFA(0.55) = 1 −(1E-10). |
| O | (0.3725, 0.55) | Undecidable pairs | INCERTITUDE — Genuine Discomfort Rate: FRR(0.55) ≈ **2.7E-4**; Imposter Discomfort Rate: FAR(0.375) ≈ **1.42E-4**. —————————————————————— Total Discomfort Rate: **4.12E-4** — DISCOMFORT / SECURITY |
| D | [0, 0.3725] | Imposter pairs | **False Reject Rate:**<br>FRR(0.3725) ≈ POFR(0.3725) = 1E-10.<br>**False Reject Safety:**<br>1−FRR(0.3725) ≈ 1−POFR(0.3725) = 1 −(1E-10). |

TABLE 3: BINARY ENCODING FOR INPUT AND FOR THE BIOMETRIC DECISION

| P | Positive Claim | "I am X" |
|---|---|---|
| N | Negative Claim | "I am not X" |

| A' | Accepted Input |
|---|---|
| R' | Rejected Input |

TABLE 4: INPUT-OUTPUT RELATION IN A FUZZY 3-VALENT DISAMBIGUATED MODEL OF IRIS RECOGNITION

| (1) Input | (2) Input encoding | (3) Similarity score | (4) Fuzzy/Modal encoding | (5) Output | (6) Output encoding |
|---|---|---|---|---|---|
| Positive / Negative claim | P / N | $p \leq t \leq 1$ | I | Accepted P, Rejected N | PA'&NR' |
| | | $n < t < p$ | O | Rejected P, Rejected N | PR'&NR' |
| | | $0 \leq t \leq n$ | D | Rejected P, Accepted N | PR'&NA' |

- $P(\cdot,\cdot)$ and $S(\cdot,\cdot)$ are defined in Table 6 and in column [c] of Table 5.

The table of additive operation S (supremum) of the Boolean algebra $(Z_8, P, S, N)$ can be read also from Fig. 4.a, if it is taken into account that for each pair $(\hat{a}, \hat{b})$ of modulo 8 integers:
- Or $\hat{a} = \hat{b}$ and then $S(\hat{a}, \hat{a}) = \hat{a}$,
- Or $\hat{a}$ and $\hat{b}$ are comparable in the partial order of the Boolean algebra $(Z_8, P, S, N)$ and then $S(\hat{a}, \hat{b}) = \hat{c} \Leftrightarrow \hat{c} = \max(\hat{a}, \hat{b})$,
- Or $\hat{a}$ and $\hat{b}$ are not comparable in the partial order of the Boolean algebra $(Z_8, P, S, N)$ and then their 'sum' is the first (the lowest) common successor, i.e. $S(\hat{a}, \hat{b}) = \hat{c} \Leftrightarrow \hat{c} = \min\{\hat{x} \in Z_8 \mid \hat{a} \leq \hat{x}, \hat{b} \leq \hat{x}\}$.

The table of multiplicative operation P (infimum) of the Boolean algebra $(Z_8, P, S, N)$ can be read also, from Fig. 4.a, if it is taken into account that for each pair $(\hat{a}, \hat{b})$ of modulo 8 integers:
- Or $\hat{a} = \hat{b}$ and then $P(\hat{a}, \hat{a}) = \hat{a}$,
- Or $\hat{a}$ and $\hat{b}$ are comparable in the purpose of partial order of the Boolean algebra $(Z_8, P, S, N)$ and then: $P(\hat{a}, \hat{b}) = \hat{c} \Leftrightarrow \hat{c} = \min(\hat{a}, \hat{b})$,
- Or $\hat{a}$ and $\hat{b}$ are not comparable in the partial order of the Boolean algebra $(Z_8, P, S, N)$ and then their

product is the last (the highest) common predecessor, i.e.:

$$P(\hat{a}, \hat{b}) = \hat{c} \Leftrightarrow \hat{c} = \max\{\hat{x} \in Z_8 | \hat{a} \geq \hat{x}, \hat{b} \geq \hat{x}\}.$$

The totally ordered subsets of partially ordered algebra $(Z_8, P, S, N)$ are represented in Fig. 4.d against two coordinates of entropy: the vertical coordinate encodes entropy as natural (arithmetic) order of modulo 8 integers, whereas the horizontal coordinate encodes what we called the absolute entropy of modulo 8 integers with respect to the product operation within the Boolean algebra $(Z_8, P, S, N)$. The absolute entropy of an element 'a' with respect to an operation 'P' within a Boolean algebra B is defined here as the number o distinct elements of the set $\{P(a, b) | b \in B\}$:

$$E(a) = card(unique(\{P(a, b) | b \in B\})).$$

Analyzing the order in the Boolean algebra $(Z_8, P, S, N)$ helped us figuring that the table of product operation is block-recursive (in three steps with blocks of dimension 1, 2, and 4), fact which further allowed the determination of an explicit formula for product calculus:

$$\forall \hat{a}, \hat{b} \in \{\hat{0}, \hat{1}, \hat{2}, \hat{3}, \hat{4}, \hat{5}, \hat{6}, \hat{7}\}:$$

$$P(\hat{a}, \hat{b}) = \hat{c} \Leftrightarrow c = \sum_{n=0}^{2} 2^n (aM2^{n+1} \geq 2^n)(bM2^{n+1} \geq 2^n),$$

where M stands for modulo and the operator '$\geq$' is considered to be a logico-arithmetical operator which returns logical values as natural numbers 0 and 1. The explicit formula of the sum operation calculus has been further defined through complementarity:

$$\forall \hat{a}, \hat{b} \in \{\hat{0}, \hat{1}, \hat{2}, \hat{3}, \hat{4}, \hat{5}, \hat{6}, \hat{7}\}:$$

$$S(\hat{a}, \hat{b}) = N(P(N(\hat{a}), N(\hat{b}))),$$

and verified against data within Table 6 and Fig. 4.d.

At this stage the following question appears: *in what formal language are well-formed the strings that define the product and the sum within the Boolean algebra $(Z_8, P, S, N)$?* They are well-defined in a formal language obtained by overloading Peano Arithmetic with the first degree logic of the propositions about the natural order between modulo 8-integers ('a $\geq$ b', 'a $\leq$ b'). In this language, the expression $(aM2^{n+1} \geq 2^n)$ which interferes in the calculation of P and S returns the natural value 0 or 1 accordingly with the true value associated to the inequality. Hence, it has been illustrated that, in order to describe the 8-valent fuzzy logic of iris recognition, Peano Arithmetic must be extended with logical support. Of course, we could see this in a reversed perspective: since it is normal that the arithmetic to describe the Boolean algebras generated by finite subsets of natural numbers, it is also normal to consider that the study of a fuzzy logic model of iris recognition has lead us to an improved model of arithmetic. However, this paper is not concerned with establishing these pure theoretical aspects. If the arithmetic should or shouldn't be overloaded with logical support, it is a question for theoreticians. Here in this paper the overloaded model of arithmetic is called Peano-2 Arithmetic and it is used to compute the operations within 8-valent Boolean algebra of iris recognition $(Z_8, P, S, N)$.

TABLE 5: ENCODING THE OUTPUT IN A FUZZY 3-VALENT DISAMBIGUATED MODEL (ISOMORPHIC REPRESENTATIONS OF THE BOOLEAN ALGEBRA $I^3$)

| [a] Symbolic encoding | [b] Binary labels | [c] Octal labels | [d] Octal labels | [e] Binary labels | [f] Meaning |
|---|---|---|---|---|---|
| IOD | 111 | 7 | 7 | 111 | PR'| NR' |
| OD | 011 | 3 | 6 | 110 | PR'| NA' |
| IO | 110 | 6 | 5 | 101 | PA'| NR' |
| ID | 101 | 5 | 4 | 100 | PA'| NA' |
| O | 010 | 2 | 3 | 011 | PR' & NR' |
| D | 001 | 1 | 2 | 010 | PR' & NA' |
| I | 100 | 4 | 1 | 001 | PA' & NR' |
| E | 000 | 0 | 0 | 000 | PA' & NA' |

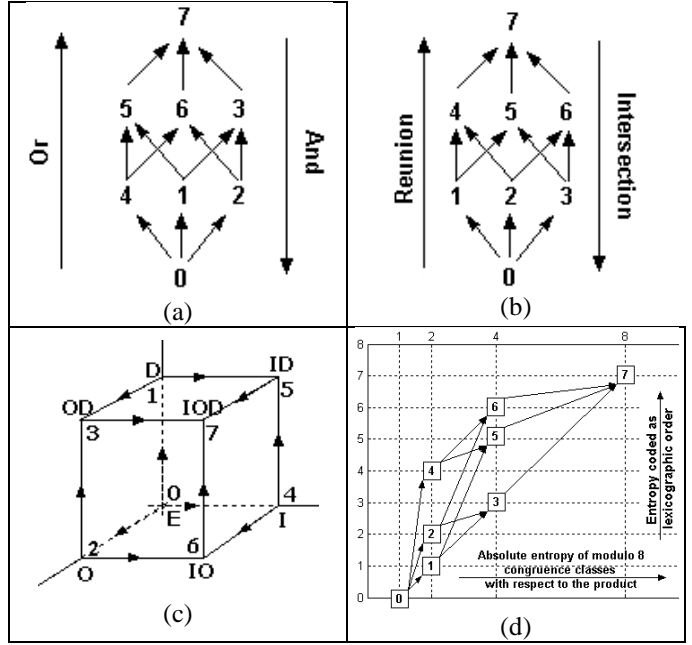

Fig. 4. Isomorphic representations of the Boolean algebra $S^3$: (a) - Boolean algebra $B^3$; (b) - Boolean algebra $I^3$; (c) - Boolean algebra $V^3$; (d) - Boolean algebra $(Z_8, P, S, N)$.

TABLE 6: PRODUCT AND SUM WITHIN BOOLEAN ALGEBRA $(Z_8, P, S, N)$

| S | 0 | 1 | 2 | 3 | 4 | 5 | 6 | 7 | E |   | P | 0 | 1 | 2 | 3 | 4 | 5 | 6 | 7 | E |
|---|---|---|---|---|---|---|---|---|---|---|---|---|---|---|---|---|---|---|---|---|
| 0 | 0 | 1 | 2 | 3 | 4 | 5 | 6 | 7 | 8 |   | 0 | 0 | 0 | 0 | 0 | 0 | 0 | 0 | 0 | 1 |
| 1 | 1 | 1 | 3 | 3 | 5 | 5 | 7 | 7 | 4 |   | 1 | 0 | 1 | 0 | 1 | 0 | 1 | 0 | 1 | 2 |
| 2 | 2 | 3 | 2 | 3 | 6 | 7 | 6 | 7 | 4 |   | 2 | 0 | 0 | 2 | 2 | 0 | 0 | 2 | 2 | 2 |
| 3 | 3 | 3 | 3 | 3 | 7 | 7 | 7 | 7 | 2 |   | 3 | 0 | 1 | 2 | 3 | 0 | 1 | 2 | 3 | 4 |
| 4 | 4 | 5 | 6 | 7 | 4 | 5 | 6 | 7 | 4 |   | 4 | 0 | 0 | 0 | 0 | 4 | 4 | 4 | 4 | 2 |
| 5 | 5 | 5 | 7 | 7 | 5 | 5 | 7 | 7 | 2 |   | 5 | 0 | 1 | 0 | 1 | 4 | 5 | 4 | 5 | 4 |
| 6 | 6 | 7 | 6 | 7 | 6 | 7 | 6 | 7 | 2 |   | 6 | 0 | 0 | 2 | 2 | 4 | 4 | 6 | 6 | 4 |
| 7 | 7 | 7 | 7 | 7 | 7 | 7 | 7 | 7 | 1 |   | 7 | 0 | 1 | 2 | 3 | 4 | 5 | 6 | 7 | 8 |
| (a) |   |   |   |   |   |   |   |   |   |   | (b) |

IV. BACK TO THE IRIS RECOGNITION PRACTICE

First of all, the artificial understanding of the experimental data obtained in iris recognition tests illustrated in Fig. 3 is expressed in the following theorem (N. Popescu-Bodorin, V.E. Balas, [8]):

*Theorem 1:* The correspondence Ψ

| Ψ: | E | D | O | I | OD | ID | IO | IOD |
|---|---|---|---|---|---|---|---|---|
| | 0 | 1 | 2 | 4 | 3 | 5 | 6 | 7 |

achieves the defuzzification of the fuzzy sets I, D and O as the elements 4, 1 and 2 from the 8-valent Boolean algebra $(Z_8, P, S, N)$, where:

$$\forall \hat{a}, \hat{b} \in \{\hat{0}, \hat{1}, \hat{2}, \hat{3}, \hat{4}, \hat{5}, \hat{6}, \hat{7}\}:$$
$$N(\hat{a}) = \hat{b} \Leftrightarrow b + a = 7,$$
$$P(\hat{a}, \hat{b}) = \hat{c} \Leftrightarrow c = \sum_{n=0}^{2} 2^n (aM2^{n+1} \geq 2^n)(bM2^{n+1} \geq 2^n),$$
$$S(\hat{a}, \hat{b}) = N\left(P(N(\hat{a}), N(\hat{b}))\right),$$

and 'M' stands for modulo.

Secondly, the artificial understanding of the experimental data obtained in iris recognition tests illustrated in Fig. 3 is logically consistent and reflected in the following theorem:

*Theorem 2:* System structure and consistency of the fuzzy 3-valent disambiguated model of iris recognition (N. Popescu-Bodorin):

Let ICP a set of iris code pairs fuzzy assigned to the fuzzy sets D, O and I by a recognition function f-R as in Fig. 3.b and Table 4. Let EICP be the set of enrollable iris code pairs, EICP = f-R$^{-1}$(I) ∪ f-R$^{-1}$(D), i.e. the support of the fuzzy concepts I and D as they appear through the fuzzy recognition function f-R, and let f-K = (EICP, f-R, {I, D}) the fuzzy formal theory of iris recognition defined over the vocabulary of enrollable iris code pairs (a restriction of the fuzzy 3-valent disambiguated model to the vocabulary of enrollable pairs).

Then the fuzzy theory f-K is f-consistent as a theory of recognition (i.e. its defuzzified form is a consistent theory of recognition).

*Proof*:

Let us consider the following partitioning of the input space:

| Input (octal) | Partitioning (octal) | State (octal) | Output (octal) | Output (binary) |
|---|---|---|---|---|
| ICP ≡ 7<br>IOD ≡ 7 | EICP ≡ 5 | D ≡ 1 | 1 | 0 |
| | | I ≡ 4 | 4 | 1 |
| | f-K theory | | | |
| | K theory | | | |
| | UICP ≡ 2 | O ≡ 2 | | |
| **Fuzzy 3-valent disambiguated model of iris recognition** | | | | |

P(4,1) = 0, and S(4,1) = 5, or equivalently, the fuzzy sets 4 and 1 are mutually exclusive and complementary to each other in the output space, which in its turn generates a subalgebra ({0,4,1,5}, P, S, N) of $(Z_8, P, S, N)$. Consequently, the fuzzy theory f-K = (EICP, f-R, {I, D}) illustrated in Fig. 3.b can by defuzzified as a crisp theory K = (EICP, R, {1, 0}) like that illustrated in Fig. 1.a. In other words, in the vocabulary of f-K theory there is no support for the concept of wolf-lamb [12] pair (there is no support for impersonation). □

V. CONCLUSION

Maintaining consistency of a biometric system is a matter of partitioning the input space into two classes: enrollable and unenrollable pairs. Consistent enrollment is mandatory in order to preserve system consistency.

The fact that the fuzzy 3-valent disambiguated model of iris recognition is incomplete (there are, indeed, undecidable pairs in the input space) is reflected in the *user discomfort* (undecidable pairs must be discarded and the user must repeat the authentication attempt) but also in the *system safety*.

As a theory of recognition, f-K is f-consistent and complete: for any pair of its vocabulary there is only one biometric decision to be given, specifically the correct one.

Consistency in iris recognition is not achievable just by setting a threshold and doing one-to-one comparisons. It can be guaranteed only by establishing a safety band (f-EER interval) and practicing one-to-all comparisons for each authentication / identification / enrollment attempt.

ACKNOWLEDGMENT

The authors would like to thank Professor Donald Monro (Dept. of Electronic and Electrical Engineering, University of Bath, UK) for granting the access to the Bath University Iris Image Database.